\documentclass[journal]{IEEEtran}

\usepackage{cite}
\usepackage{amsmath,amssymb,amsfonts}
\usepackage{algorithmic}
\usepackage{textcomp}
\usepackage{hyperref}
\usepackage{wrapfig}
\usepackage{xcolor}
\usepackage{soul}
\usepackage{gensymb}
\usepackage{adjustbox}
\usepackage{lipsum} 
\usepackage{caption}
\usepackage{subcaption}
\usepackage{multirow}
\usepackage{gensymb}
\usepackage[linesnumbered,ruled,noline]{algorithm2e}

\begin{document}

\newcommand{\datasetUrl}{\url{https://github.com/ansfl/MoRPINet}} 
\newcommand{\hlc}[2][yellow]{{%
    \colorlet{foo}{#1}%
    \sethlcolor{foo}\hl{#2}}%
}

\title{Snake-Inspired Mobile Robot Positioning with Hybrid Learning}
\author{Aviad~Etzion, Nadav~Cohen, Orzion~Levy, Zeev~Yampolsky, and~Itzik~Klein
\thanks{A. Etzion, N. Cohen, O. Levi, Z. Yampolsky and I. Klein are with the Hatter Department of Marine Technologies, University of Haifa, Israel.}
\thanks{Manuscript received November 2024}}

\maketitle
\begin{abstract}
Mobile robots are used in various fields, from deliveries to search and rescue applications. Different types of sensors are mounted on the robot to provide accurate navigation and, thus, allow successful completion of its task. In real-world scenarios, due to environmental constraints, the robot frequently relies only on its inertial sensors. Therefore, due to noises and other error terms associated with the inertial readings, the navigation solution drifts in time. To mitigate the inertial solution drift, we propose the MoRPINet framework consisting of a neural network to regress the robot's travelled distance. To this end, we require the mobile robot to maneuver in a snake-like slithering motion to encourage nonlinear behavior.
MoRPINet was evaluated using a dataset of 290 minutes of inertial recordings during field experiments and showed an improvement of 33\% in the positioning error over other state-of-the-art methods for pure inertial navigation. 
\end{abstract}


\IEEEpeerreviewmaketitle

\section{Introduction}

\IEEEPARstart{A}{} number of key factors contribute to the growth of mobile robots, including increased efficiency, productivity, and versatility.  There are many applications for mobile robots. These include fruit picking or monitoring in agriculture applications, indoor or outdoor deliveries, inspection or transportation on construction sites. They also include data collection for multiple purposes and operation in hazardous environments.
\\
There are several kinds of mobile robots 
including: legged robots, tracked slip locomotion and wheel-based mobile robots \cite{rubio2019review}. The latter are easier to design, cheaper to build, and are less complex to control. 
One of the most crucial aspect of mobile robots design is its navigation capabilities as they are responsible for determining the robot position and orientation. To that end, the robot relies on different sensors. Commonly, the positioning is performed by using sensors such as camera \cite{vision2002}, LiDAR \cite{lidar2018}, sonar \cite{Leonard2012}, global navigation satellite  system (GNSS) \cite{farrell2008aided}, inertial sensors \cite{jimenez2010imu} or odometer \cite{Odometer2021}. A nonlinear filter, such as the extended Kalman filter\cite{kalman1960new}, can also be used to combine these sensors with inertial sensors. However, in practical situations only the inertial readings are available. For example, in poor lighting conditions the vision based system are degraded, or when operating indoors the GNSS signals are unavailable.
\\
The inertial sensors are grouped in inertial measurement unit (IMU), which includes three perpendicular accelerometers and three perpendicular gyroscopes \cite{Titterton2005}. Inertial sensors are commonly used because of their simplicity of installation, small size, low-cost, and high sampling rate. 
Yet, when integrating the inertial sensor reading, the navigation solution drift over time because the noise and the other error terms.
\\
Recently, machine and deep learning approaches have demonstrated significant advancements over model-based methods in inertial sensing across various platforms \cite{Itzik2022DGON, cohen2023survey,chen2024survey}, including autonomous underwater vehicles \cite{yona2021compensating,saksvik2021auv}, quadrotors \cite{shurin2022quadnet,zhang2022dido}, and pedestrians \cite{asraf2021pdrnet,chen2020pdr}. In mobile robot navigation, deep learning is primarily utilized with vision systems, building on extensive research in neural networks for image processing. The two main methods for vision-based navigation involve comparing landmarks to a predefined map \cite{DOURADO2019859vision,kim2015probabilisticVision} or creating a real time map for simultaneous localization and mapping \cite{wang2019machineSlam,SLAM2004}.
\\
Inspired by the locomotion of snakes \cite{jayne1986kinematics,hu2009mechanics}, a novel method has been proposed that incorporates the serpentine movement of a mobile robot, along with a corresponding algorithm, to enhance motion accuracy. Snakes use serpentine slithering to compensate for their lack of legs, enabling them to move efficiently, conserve energy, and maintain excellent maneuverability even in rough terrain. By adopting this type of locomotion in mobile robots, maneuverability can be preserved, albeit with a potential increase in power consumption. Additionally, this mode of locomotion enriches sensor measurements, leading to a high signal-to-noise ratio and achieving high positional accuracy. Initially, the concept was explored with quadrotors, where Shurin et al. \cite{Shurin2020} developed a model-based method to estimate quadrotor positions and later \cite{shurin2022quadnet} improved this solution by employing a deep learning network to estimate step length and altitude. Further research demonstrated that using multiple IMUs on the quadrotor yields better results \cite{hurwitz2023quadrotor}. Recently, we proposed the Mobile Robot Pure Inertial Navigation (MoRPI) framework \cite{etzion2023morpi}, which is based on periodic movement and employs an empirical formula to determine step length, similar to the approach used with quadrotors. However, mobile robots have low amplitudes, slow motion, minimal gravity changes, and frequent periods, hence they require unique considerations that cannot be directly transferred from quadrotor and pedestrian solutions, which makes MoRPI an excellent alternative to pure inertial navigation solutions.
\\
In MoRPI, the method requires an additional calibration phase for gain calibration. The gain is sensitive to motion parameters and is a primary factor contributing to position error. Furthermore, the solution is inherently limited to peak-to-peak segments, which reduces the update rate of the positioning.
\\
Therefore, in this paper, we present a deep-learning-based algorithm incorporating serpentine dynamics. The contribution of this paper:
\begin{enumerate}
    \item Inspired by snake-like slithering motion, we present an approach for estimating distance increments for mobile robots in a situation of pure inertial navigation by using deep learning and inertial sensors. Emulating snake-like slithering motion increases the signal-to-noise ratio of the inertial sensor data and enables accurate navigation.
    \item Our dataset contains 290 minutes of ground-truth trajectories from an GNSS-RTK sensor in $10 Hz$ and STD of $0.1 m$ with recordings from about 5 different IMUs in $120 Hz$ (each IMU has 58 min). The dataset and the code are publicly available and can be found here: \datasetUrl.
\end{enumerate}
Our approach achieves better resolution by using small time windows and enhances robustness by training on a diverse dataset. As a result, it shows a 99\% improvement compared to traditional INS solutions and a 33\% improvement over MoRPI. With an update rate of 5 Hz, about 20 times faster than MoRPI. The proposed neural network is also lightweight and can be implemented on edge devices. The small amplitude suggested in this manuscript enables the robot to plan its path effectively using various sensors, such as cameras and Lidar, scan and classify the environment, and transport objects, without limiting most typical mobile robot applications. We demonstrate the effectiveness of our method through field experiments using a mobile robot equipped with RTK-GNSS and IMU.
\\
The rest of the paper is organized as follows: Section \ref{sec:MBA} presents the INS equations and the MoRPI method. Section \ref{sec:PA} describes the proposed approach. Section \ref{sec:ER} explains the experiments and gives the results, and Section \ref{sec:CON} gives the conclusions of this paper. 
\section{Model-Based Approaches} \label{sec:MBA}
This section provides a brief overview of two model-based solutions used later in comparison to our proposed approach.

\subsection{Inertial Navigation System}

An inertial navigation system (INS) provides a complete navigation solution, consisting of the position vector, velocity vector, and orientation.
The INS equations of motion are commonly expressed in the navigation frame with north-east-down coordinates \cite{Groves2013}. As mobile robot navigation is addressed,  a local coordinate frame (l-frame), is adopted. It is located at the initial position of the robot and its coordinates are in the the north-east-down directions. The position vector rate of change is
\begin{equation}
    \dot{\boldsymbol{p}}^l =\boldsymbol{v}^l\label{eq:ins_eq1}
\end{equation}
where $\boldsymbol{p}^l$ is the position vector expressed in the l-frame and $\boldsymbol{v}^{l}$ is the velocity vector expressed in the l-frame. 
\\
The velocity rate of change is:
\begin{equation}
    \dot{\boldsymbol{v}}^l = \mathbf{R}_b^l\boldsymbol{f}_{ib}^b+\boldsymbol{g}^l\label{eq:ins_eq2}
\end{equation}
\noindent where $\boldsymbol{g}^{l}$ is the gravity vector expressed in the l-frame, $\mathbf{R}^{l}_{b}$ is the transformation matrix from the body frame to the l-frame and $\boldsymbol{f}^{b}_{ib}$ is the specific force vector measured by the accelerometer and expressed in body frame. 
\\
The rate of change of the transformation matrix is given by:
\begin{equation}
    \dot{\mathbf{R}}_b^l =\mathbf{R}_b^l\mathbf{\Omega}_{ib}^b\label{eq:ins_eq3}
\end{equation}
\noindent where $\mathbf{\Omega}^{b}_{ib}$ is the skew-symmetric form of the angular rate measured by the gyroscope, expressed in the body frame. 
\\
Notice that as our scenarios include low-cost inertial sensors and short time periods, the earth turn rate and the transport rate are neglected in \eqref{eq:ins_eq2}-\eqref{eq:ins_eq3}.

\subsection{MoRPI}
The MoRPI approach is based on periodic movement and employs an empirical formula to determine the peak to peak distance \cite{etzion2023morpi}. It uses accelerometer or gyroscope readings for peak-to-peak event detection. Then, by using Weinberg's step length estimation approach, the peak-to-peak distance is estimated by:
\begin{equation}
    s=G\big(\max\left(\boldsymbol{f}^b\right)-\min\left(\boldsymbol{f}^b\right)\big)^\frac{1}{4} \label{eq:weinberg}
\end{equation}
where $s$ is the peak-to-peak distance, $G$ is the approach's gain, and $\boldsymbol{f}^b$ is the sequence of accelerometer readings between two successive peak detection.\\
Two MoRPI approaches are available:
\begin{itemize}
    \item \textbf{MoRPI-A:} Uses the accelerometer readings for peak detection. The most sensitive axis in the snake-like slithering motion is the one that is perpendicular to the motion progress-direction. Assuming the accelerometer sensitive axes directions are: x - points towards the path segment end point, y - is the perpendicular axis to x in the ground planar and z - is oriented to complete the right-hand rule with the x and y axes. Then the specific force readings in the y axis are plugged into \eqref{eq:weinberg}.
    \item \textbf{MoRPI-G:} Uses the gyroscope readings for peak detection. To cope with limited dynamics cases, for example when driving in narrow passes, the signal to noise ratio in the accelerometer readings is low. Hence, an approach that relies only on the gyroscope was suggested. Here, the z-axis angular rate, $\boldsymbol{\omega}_z$, readings are used in \eqref{eq:weinberg} instead of $\boldsymbol{f}_{y}^b$ axis of the accelerometer.
\end{itemize}
The MoRPI gain, $G$, is estimated by moving the robot at known distance prior the method can be used. Each approach requires a different gain value.
The peak-to-peak distance estimation together with the gyro-based heading \eqref{eq:ins_eq3}, and initial conditions are used to propagate the mobile robot two-dimensional position by
\begin{align}
            x_{k+1} &= x_k+s_{k}\cos \psi_k \label{eq:morpi_x}\\
            y_{k+1} &= y_k+s_{k}\sin \psi_k \label{eq:morpi_y}
\end{align}
where $x$ and $y$ are the robot's position coordinates, $\psi$ is the heading, and $k$ is the peak index.
\\
The MoRPI model assumes that the trajectory is composed of connected straight lines due to the position propagation model in \eqref{eq:morpi_x}-\eqref{eq:morpi_y} which is based on the peak to peak distance. 
\section{MoRPINet Framework} \label{sec:PA}
\begin{figure}[ht!]
\centering
\includegraphics[width=0.6\linewidth]{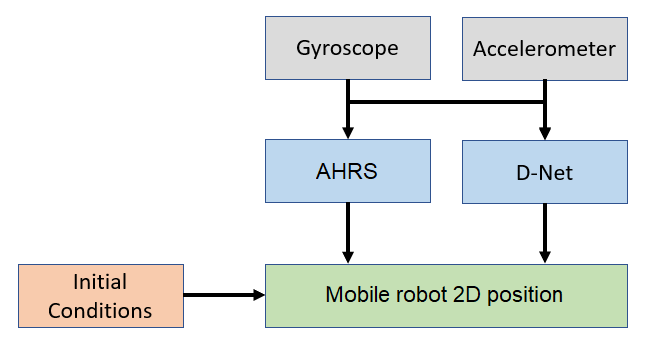}
\caption{MoRPINet is an inertial positioning approach for mobile robots.}
\label{fig:method}
\end{figure}

Mobile robots with inertial sensors mounted on them produce mostly noisy inertial readings (low signal-to-noise ratio) while travailing with a nearly constant velocity, making them unsuitable for estimating position by neural networks. In such scenarios, it is possible to use the model-based INS equations of motion \eqref{eq:ins_eq1}-\eqref{eq:ins_eq3}, but positioning errors will be high. To overcome this situation the serpentine locomotion is adopted for mobile robots movement. In its nature, the serpentine locomotion is a dynamic motion with angular velocity and linear acceleration. As a consequence, this motion is expected to enrich the inertial signals and increase the signal to noise ratio for both accelerometers and gyroscopes. This will enable more features in the data processed by neural networks enabling the extraction of relevant positioning information. To this end, we propose MoRPINet, a pure inertial positioning approach. MoRPINet requires the mobile robot to move in a serpentine dynamics and splits the trajectory reconstruction into two parts:
\begin{enumerate}
    \item \textbf{Distance Estimation:} D-Net, a neural network architecture for distance regression is proposed. It is a simple yet efficient structure consisting of a one-dimension convolution layers (1DCNN) and a fully connected (FC) head for distance estimation, based only on the inertial readings.
    \item \textbf{Heading Estimation:} The well-established 
    Madgwick filter \cite{madgwick2010efficient} is adopted for heading estimation based on the inertial readings.
\end{enumerate}
Our proposed approach, MoRPINet, is illustrated in Figure \ref{fig:method}. In the following subsections we elaborate on each part of MoRPINet.
\subsection{D-Net: Distance Estimation Network}
\begin{figure}[ht!]
\centering
\includegraphics[width=0.6\linewidth]{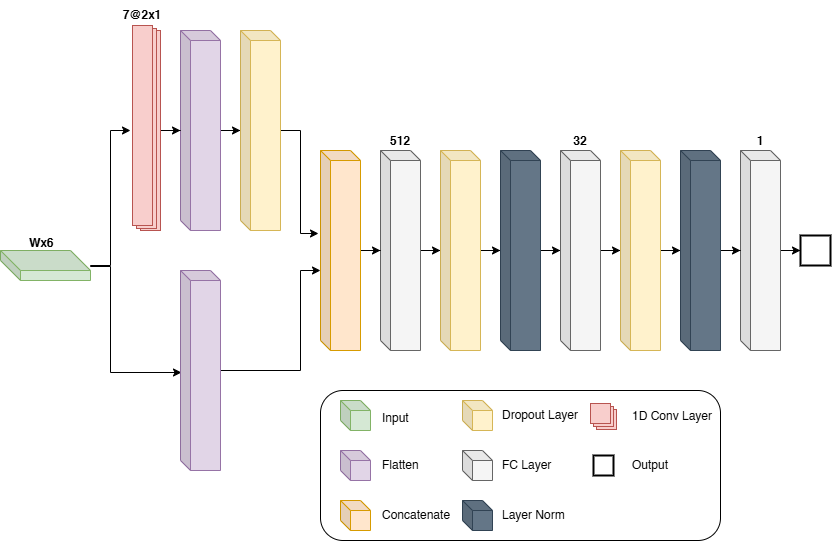}
\caption{D-Net: Our inertial-based distance estimation network architecture.}
\label{fig:cnn}
\end{figure}

The network architecture of the proposed approach is based on 1DCNN and FC layers. The network receives a time window with \textit{n} samples of data from the three-axis gyroscope and three-axis accelerometer. Initially, the input data is processed through a convolutional layer comprising of seven filters, each with a size of 2x1. Subsequently, the extracted features are flattened, and dropout is applied to prevent overfitting. The input is then flattened again and concatenated with the output of the dropout layer. This data is fed through two FC layers with 512 and 32 neurons, each followed by dropout and layer normalization. Both the convolutional and fully connected layers utilize ReLU activation \cite{NIPS2012relu} functions to introduce nonlinearities into the model. An illustration of the architecture is given in Figure \ref{fig:cnn}. The output of the network is a regressed value for the travelled distance at the given time window.

\subsection{MoRPINet Training Process} \label{subsec:TP}
In the process of training a deep learning network, the main objective is to determine the weights and biases that solve the given problem. By assuming a $m_{1} \times m_{2}$ filter (or kernel) the output of the convolutional layer can be written as follows \cite{bengio2017deep}: 
\begin{equation}\label{eq:conv}
    \centering    \mathbf{C}_{\dot{\imath}\dot{\jmath}}^{(\ell)}=\sum_{\alpha=0}^{m_{1}}\sum_{\beta=0}^{m_{2}}\boldsymbol{\omega}_{\alpha\beta}^{(r)}\boldsymbol{a}_{(\dot{\imath}+\alpha)(\dot{\jmath}+\beta)}^{(\ell-1)}+\boldsymbol{b}^{(r)}
\end{equation}
where $\boldsymbol{\omega}_{\alpha\beta}^{(r)}$ is the weight in the $(\alpha,\beta)$ position of the $r^{th}$ convolutional layer, $\boldsymbol{b}^{(r)}$ represents the bias of the $r^{th}$ convolutional layer, and $\boldsymbol{a}_{\dot{\imath}\dot{\jmath}}^{(\ell-1)}$ is the output of the preceding layer.
\\
The fully-connected layers are built by a number of neurons. The following equation expresses the output of each neuron:
\begin{equation}\label{eq:fc}
    \centering
        \boldsymbol{z}_{\dot{\imath}}^{(\ell)}=\sum_{\dot{\jmath}=1}^{n_{\ell-1}}\boldsymbol{\omega}_{\dot{\imath}\dot{\jmath}}^{(\ell)}\boldsymbol{a}_{\dot{\jmath}}^{(\ell-1)}+\boldsymbol{b}_{\dot{\imath}}^{(\ell)}
\end{equation}
 where $\boldsymbol{\omega}_{\dot{\imath}\dot{\jmath}}^{(\ell)}$ is the weight of the $\dot{\imath}^{th}$ neuron in the $\ell^{th}$ layer associated with the output of the $\dot{\jmath}^{th}$ neuron in the $(\ell-1)^{th}$ layer, $\boldsymbol{a}_{\dot{\jmath}}^{(\ell-1)}$, $\boldsymbol{b}_{\dot{\imath}}^{(\ell)}$ represents the bias in layer $\ell$ of the $\dot{\imath}^{th}$ neuron, and $n_{\ell-1}$ represents the number of neurons in the $\ell-1$ layer.
\\
Equation \eqref{eq:fc} represents a linear process. Therefore, for the network to cope with nonlinear problems, the neuron's output $ \boldsymbol{z}_{\dot{\imath}}^{(\ell)}$ has to go through a nonlinear activation function, $h(\cdot)$, which results in \cite{gonzalez2018deep}:  
\begin{equation}\label{eq:nl}
    \centering
        \boldsymbol{a}_{\dot{\imath}}^{(\ell)}=h(\boldsymbol{z}_{\dot{\imath}}^{(\ell)}).
\end{equation}
\\
Specifically, we employ, for all the layers in the model, the rectified linear unit (ReLU) \cite{agarap2018deep} as our nonlinear activation function $h(\cdot)$, which has a strong mathematical and biological basis. The ReLU activation function is defined by 
    \begin{equation}\label{eq:relu}
    \centering
        ReLU(\boldsymbol{z}_{\dot{\imath}}^{(\ell)})=max(0,\boldsymbol{z}_{\dot{\imath}}^{(\ell)}).
   \end{equation}
\\
The mean absolute error (MAE) loss function is used for the forecasting process:
\begin{equation}\label{eq:loss}
    \centering
        L(\boldsymbol{y}_{\dot\imath},\hat{\boldsymbol{y}}_{\dot\imath})=\frac{1}{n}\mid\mid \boldsymbol{y}_{\dot\imath}-\hat{\boldsymbol{y}}_{\dot\imath}\mid\mid
\end{equation}
where $\boldsymbol{y}_{\dot\imath}$ is the ground truth (GT) distance, $\hat{\boldsymbol{y}}_{\dot\imath}$ is the predicted value of the distance and \textit{n} is the number of items in the batch. In order to generate the prediction, the input has to go through \eqref{eq:conv}-\eqref{eq:loss}, in a process called the forward propagation \cite{zhao2017convolutional}.
The learning process is performed by a stochastic gradient descent method, where the weights and biases are update by
\begin{equation}\label{eq:gds}
    \centering
        \boldsymbol{\theta}=\boldsymbol{\theta}-\eta\nabla_{\theta}J({\boldsymbol{\theta}})\; , \quad  \boldsymbol{\theta}=[\boldsymbol{\omega}\quad \boldsymbol{b}]^{T}
\end{equation}
Here we used $J(\boldsymbol{\theta})$ to present the loss function to emphasize that the parameter now is the vector $\boldsymbol{\theta}$, where $\boldsymbol{\theta}$ is the vector of weights and biases. $\eta$ is the learning rate, and $\nabla_{\theta}$ is the gradient operator.
\\
In the suggested approach, the adaptive moment estimation (Adam) optimization algorithm \cite{bock2019proof} was employed for better convergence of D-Net. In addition, we used learning rate reduction on plateau as scheduler with a factor of 0.5.
\\
The number of epochs used in the training was 300 with a batch size of 2048. The initial learning rate was set to 0.0025. The dropout probability for the first dropout layer (applied after flattening) was set to 0.1, while for the two subsequent layers following the FC layers, it was 0.5.
\\
For the training dataset we used windows size of $W=24$ accelerometer and gyroscope samples ($6 \times 24$) with overlap of 12 samples between two successive time windows.
\subsection{Madgwick Filter} \label{subsec:MF}
We employ the Madgwick filter \cite{madgwick2010efficient} for heading determination, due to its low computational load, high rate solution and the ability to generalise the filter to include magnetometer measurements for future work. The Madgwick filter has two main parts:
\begin{enumerate}
    \item \label{itm:ori_gyro} \textbf{Orientation from angular rate:} The orientation is computed by numerically integrating the angular rates measurements from the three-axis gyroscope. First, a quaternion containing the measurement is defined as:
\begin{equation}
    \boldsymbol{\omega}_q = \begin{bmatrix}
       0 &
       \omega_x &
       \omega_y &
       \omega_z
   \end{bmatrix} \\
\end{equation}
where $\boldsymbol{\omega}_q$ is a quaternion and $\omega_i, i=x,y,z$ is the angular rate components, as measured by the gyroscope.
The quaternion rate of change is
\begin{equation}
     \dot{\boldsymbol{q}}_{w,t} = \frac{1}{2} \hat{\boldsymbol{q}}_{t-1} \cdot \boldsymbol{\omega}_q
\end{equation}
where $\boldsymbol{\hat{q}}_{t-1}$ is the estimated quaternion at time $t-1$ and $\boldsymbol{\dot{q}}_{w,t}$ is the quaternion rate of change.
\\
Finally, the updated quaternion is
\begin{equation} \label{eq:madgwick_gyro}
     \boldsymbol{q}_{w,t} =  \hat{\boldsymbol{q}}_{t-1} +  \dot{\boldsymbol{q}}_{w,t} \Delta {t}
\end{equation}
where $\boldsymbol{q}_{w,t}$ is the quaternion integrated orientation solution and $\Delta t$ is the sampling period.

\item \label{itm:ori_acc} \textbf{Orientation from vector observations:} By using the gradient descent algorithm the error between the rotated gravitational field and the accelerometer measurements can be minimized, and later used to update the gyroscope-based quaternion \eqref{eq:madgwick_gyro}. The normalized gravitational field in the initial local coordinate frame is
\begin{equation}
    \boldsymbol{g}_e = \begin{bmatrix}
       0 &
       0 &
       0 &
       1
   \end{bmatrix} \\
\end{equation}
The accelerometer measurement is normalized so that the sum of the absolute values of its components equals one. The normalized reading expressed in the body frame is
\begin{equation}
    \boldsymbol{f}_b = \begin{bmatrix}
       0 &
       f_x &
       f_y &
       f_z
   \end{bmatrix} \\
\end{equation}
where $f_i, i=x,y,z$ is the normalized specific force components. Ideally, we want to find orientation such that the rotated gravitational field will be as close as to the accelerometer measurements. Therefore, the objective function is
\begin{equation}
    \boldsymbol{f}\left(\boldsymbol{\hat{q}},\boldsymbol{g}_e, \boldsymbol{f}_b \right) = \boldsymbol{\hat{q}}^* \cdot \boldsymbol{g}_e \cdot \boldsymbol{\hat{q}} - \boldsymbol{f}_b
\end{equation}
where $\hat{q}$ is the orientation in quaternion and $\hat{q}^*$ is its conjugate number.
Then, the gradient of the objective function with the last estimated quaternion is used to update the quatrenion recursively:
\begin{equation} \label{eq:madgwick_acc}
    \boldsymbol{q}_{\nabla f,t} = \hat{\boldsymbol{q}}_{t-1} - \mu \frac{\nabla\boldsymbol{f}}{\|\nabla\boldsymbol{f}\|}
\end{equation}
where $\boldsymbol{q}_{\nabla f, t}$ is the estimated orientation quaternion calculated by the gradient of the objective function at time $t$ and $\mu$ is the convergence rate of $\boldsymbol{q}_{\nabla f, t}$.
\end{enumerate}
The final solution of the filter is done by weighted fusion between \eqref{eq:madgwick_gyro} and \eqref{eq:madgwick_acc} resulting with:
\begin{equation} \label{eq:madgwick_final}
    \hat{\boldsymbol{q}}_t = \gamma \boldsymbol{q}_{\nabla f,t} + \left(1-\gamma \right) \boldsymbol{q}_{w,t}
\end{equation}
with $0\leq \gamma \leq 1$.
\\
The conversion to the heading angle is \cite{Titterton2005}:
\begin{equation} \label{eq:psi_ahrs}
    \psi_{AHRS} = \arctan\left(\frac{2\left(q_2 q_3 + q_1 q_4\right)}{q_1^2 + q_2^2 - q_3^2 - q_4^2} \right)   
\end{equation}
where $q \triangleq \begin{bmatrix}
    q_1 & q_2 & q_3 & q_4
\end{bmatrix}$ is the quaternion solution from \eqref{eq:madgwick_final} and $\psi_{AHRS}$ is the yaw angle extracted from it.

\subsection{Summary} \label{subsec:MP}
Once the distance and heading are estimated, the dead reckoning position update equations are
\begin{align}
    x_{k+1} &= x_k + s_{Dnet,k}\cos \overline{\psi}_k \label{eq:morpinet_x}\\
    y_{k+1} &= y_k + s_{Dnet,k}\sin \overline{\psi}_k \label{eq:morpinet_y}
\end{align}
where $s_{Dnet,k}$ is the D-Net estimated distance at time k. The heading angle is estimated in the inertial sensors sampling rate which is faster than the distance estimation rate (window size).  Also, as the window size is short (part of a second) the robot dynamics changes slowly during that period. As a result,  we use the average heading angle in the dead-reckoning equations \eqref{eq:morpinet_x}-\eqref{eq:morpinet_y}. The average heading angle is defined by:
\begin{equation} \label{eq:psi_morpinet}
    \overline{\psi}_k = \frac{1}{W} \sum_{i=0}^W \psi_{AHRS}\left(t_{k,i}\right)
\end{equation}
where $\psi_{AHRS} \left(t_{k,i} \right)$ is the extracted yaw angle from the Madgwick filter \eqref{eq:psi_ahrs} at window k and sample i.
\\
In contrast with MoRPI position propagation \eqref{eq:morpi_x}-\eqref{eq:morpi_y}, MoRPINet position propagation depends on a fixed time window and not on  the unknown varying peak to peak time. This is also one of the benefits of MoRPINet as, generally, a shorter window size reduces the dead-reckoning propagation error.
\\
To summarize, MoRPINet framework has three phases:
\begin{enumerate}
    \item \textbf{Distance Estimation:} A shallow, yet efficient, network to estimate the distance over the required time window (Section \ref{subsec:TP}).
    \item \textbf{Heading determination:} Madgwick filter is employed for heading estimation \eqref{eq:psi_ahrs} in the required window size (Section \ref{subsec:MF}).
    \item \textbf{Position Update:} A dead-reckoning position update \eqref{eq:morpinet_x}-\eqref{eq:morpinet_y} is applied based on the distance and heading from previous phases.
\end{enumerate}
The MoRPINet  algorithm  is presented in Algorithm \ref{alg:morpinet}.

\begin{algorithm}[!ht]
\SetAlgoLined
\KwData{$\boldsymbol{p}_0$, $\psi_0$, $\left\{\boldsymbol{f^b}, \boldsymbol{\omega^b} \right\}$}
$k \gets 1$\;
$\boldsymbol{p}_k \gets \boldsymbol{p}_0$\;
$\overline{\psi}_{k-1} \gets \psi_0$\;
\While{True}{
$s_{Dnet,k} \gets Dnet\left(\left\{\boldsymbol{f^b_i}, \boldsymbol{\omega^b_i}\right\}_{i=t_k,\dots,t_{k+1}} \right)$\;
$\boldsymbol{\psi}_k^{AHRS} \gets Madgwick\left(\left\{\boldsymbol{f^b_i}, \boldsymbol{\omega^b_i}\right\}_{i=t_k,\dots,t_{k+1}} \right)$\;
$\overline{\psi}_k \gets mean\left(\boldsymbol{\psi}_k^{AHRS} \right)$\;
$\boldsymbol{p}_k \gets update\_position\left(s_{Dnet,k}, \overline{\psi}_k \right)$\;
$k \gets k+1$\;
}
\caption{MoRPINet Algorithm}\label{alg:morpinet}
\end{algorithm}

\section{Experiment Setup and Dataset} \label{sec:ER}
\subsection{Experiment Setup}
\begin{figure}[ht!]
\centering
\includegraphics[width=8cm]{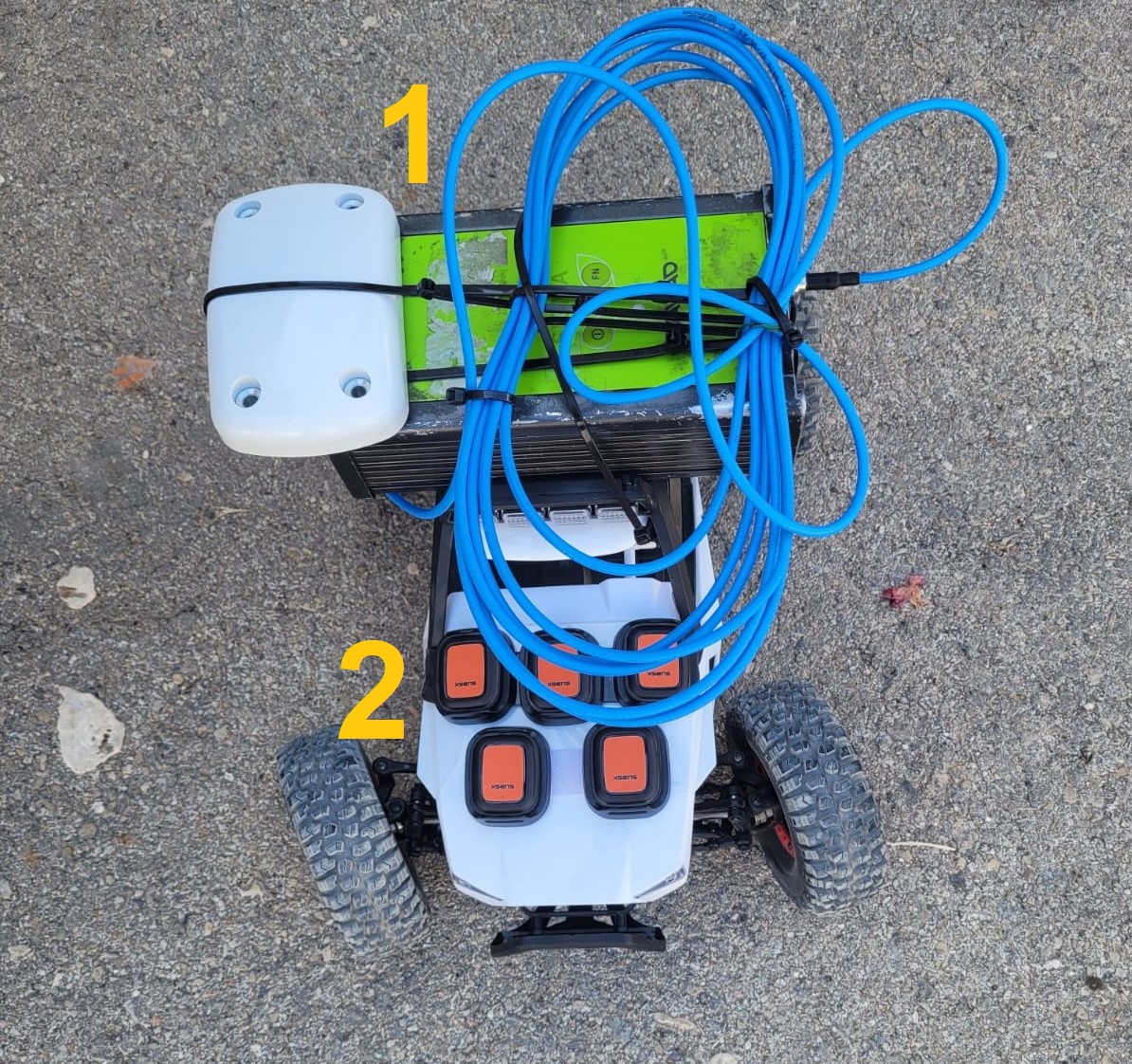}
\caption{Our RC car mounted with (1) GNSS RTK receiver and (2) five IMU sensors.}
\label{fig:car}
\end{figure}

A remote control (RC) car was used to conduct the experiments. The car model, STORM Electric 4WD Climbing Car, has dimensions of $385 \times 260 \times 205 mm$, with a wheelbase of $253 mm$ and a tire diameter of $110 mm$. The RC car was equipped with a Javad SIGMA-3N RTK sensor, which provides positioning measurements with an accuracy of $10 cm$ at a sample rate of $10$Hz, serving as the GT \cite{Javad}. Additionally, five IMUs were mounted on a rigid surface at the front of the RC car. The experimental setup is presented in Figure \ref{fig:car}. We worked with the Movella DOT IMUs, capable of operating at a $120$Hz \cite{Xsens}. The DOT software allows synchronization between the IMUs. The associated noise and bias values of the accelerometer and gyroscope are presented in Table \ref{table:1}.     

\begin{table}[ht!]
\caption{Movella DOT IMU sensor specifications. Both sensors measure at sampling rate of 120Hz.}
\centering
\begin{tabular}{|cc|cc|}
\hline
\multicolumn{2}{|c|}{Gyro} & \multicolumn{2}{c|}{Accelerometer}\\ 
\hline
\multicolumn{1}{|c|}{Bias [$\degree/h$]} & Noise [$\degree/s/\sqrt{Hz}$] & \multicolumn{1}{c|}{Bias [mg]} & Noise [$\mu{g}/\sqrt{Hz}$] \\ 
\hline
\multicolumn{1}{|c|}{10} & 0.007 & \multicolumn{1}{c|}{0.03} & 120\\ 
\hline
\end{tabular}
\label{table:1}
\end{table}

\subsection{Dataset} \label{subsec:DS}
\begin{figure*}[htb!]
    \begin{subfigure}[t]{0.5\linewidth}
    \includegraphics[width=\linewidth]{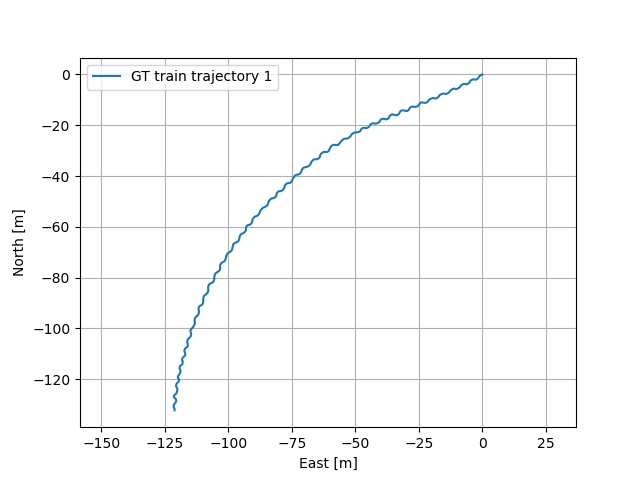}
        \caption{Train trajectory no. 3}
    \end{subfigure}\hfill
    \begin{subfigure}[t]{0.5\linewidth}
    \includegraphics[width=\linewidth]{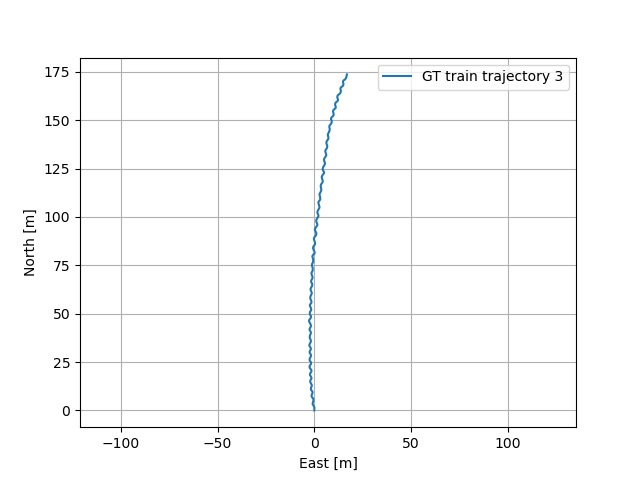}
        \caption{Train trajectory no. 4}
    \end{subfigure}
    
    \begin{subfigure}[b]{0.5\linewidth}
    \includegraphics[width=\linewidth]{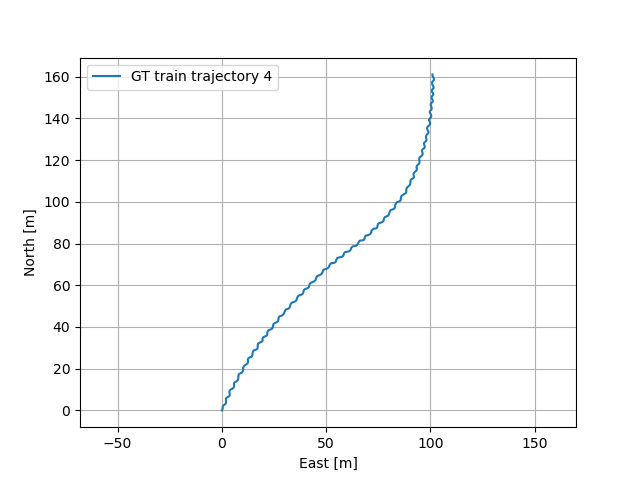}
        \caption{Train trajectory no. 5}
    \end{subfigure}\hfill
    \begin{subfigure}[b]{0.5\linewidth}
        \includegraphics[width=\linewidth]{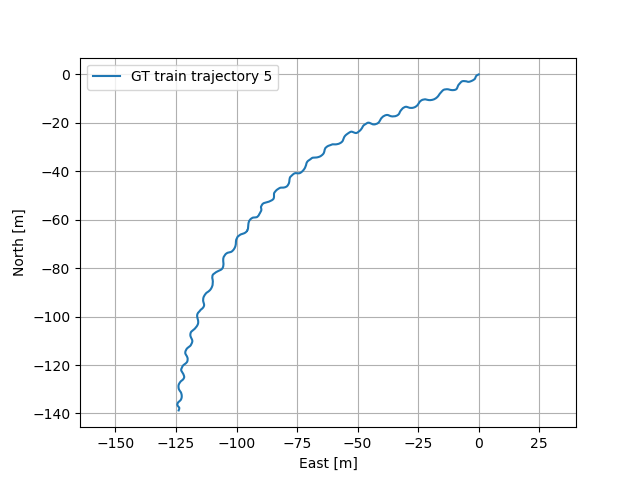}
        \caption{Train trajectory no. 6}
    \end{subfigure}
    \caption{Example of four different trajectories. The mobile robot is moving in a snake-like slithering motion.}
    \label{fig:gt}
    \end{figure*}

\begin{figure}[ht!]
\centering
\includegraphics[width=0.6\linewidth]{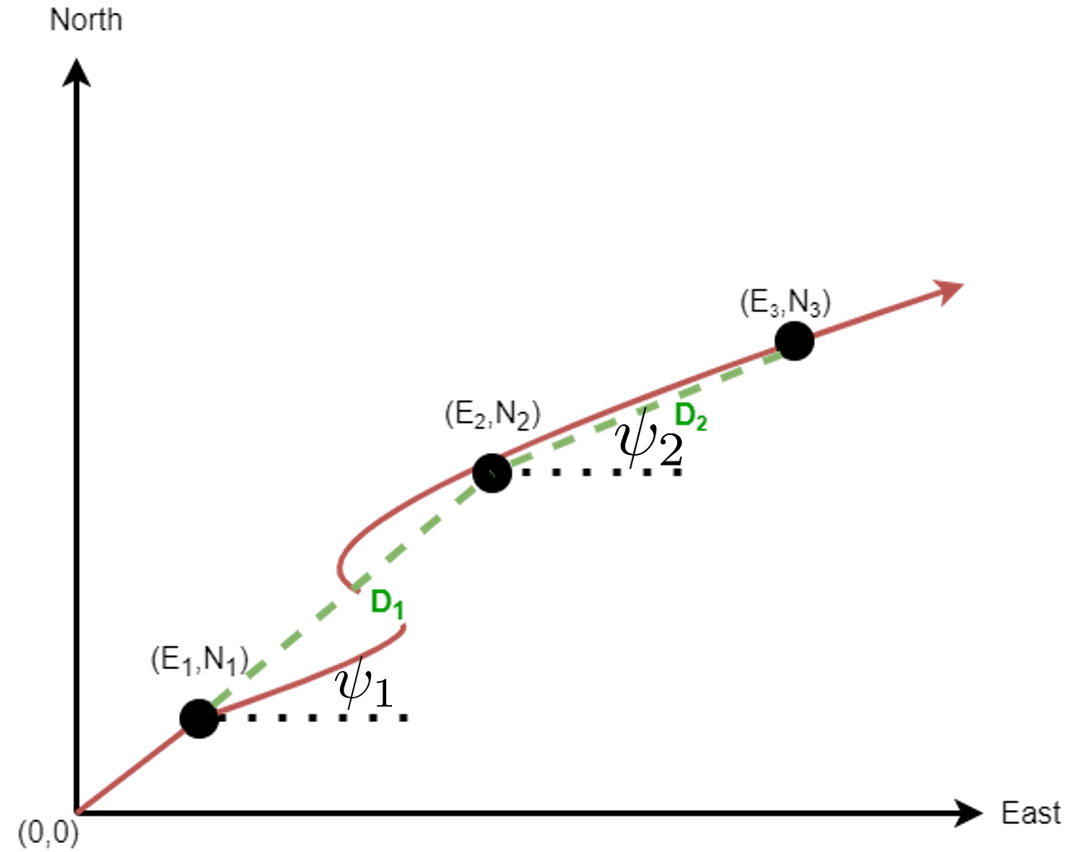}
\caption{An illustration of ground-truth trajectory (in red). The distance (in green) and the heading angles are calculated using \eqref{eq:gt_dist}-\eqref{eq:gt_angs}.}
\label{fig:calc}
\end{figure}

Thirteen distinct trajectories were recorded during field experiments with a total of 58 minutes for a single IMU and  290 for the entire dataset. Each trajectory includes GT data obtained from the GNSS RTK, and inertial measurements recorded simultaneously by the five IMUs mounted on the RC car.
\\
Each recording session began with a one-minute static period, which was utilized for stationary calibration and synchronize the timing between the IMU and GNSS RTK measurements. Synchronization between the two sensors was achieved during post-processing. The biases of the IMU's accelerometers and gyroscopes were determined by averaging the IMU measurements for a few seconds taken during the stationary period of each recording. They were subsequently offset from the entire dataset as needed. To this end, we assumed that the IMU is parallel to the ground with almost zero roll and pitch angles, allowing for accelerometer calibration. 
Figure \ref{fig:gt} shows GNSS-RTK position measurements of four different snake-like slithering motion trajectories.
\\
\\
The recorded trajectories were divided into train and test datasets:
\begin{enumerate}
\item \textbf{Train Dataset}:
Seven trajectories are used for the neural network training dataset. All seven include snake-like slithering motion with variable frequency and amplitude with slowly changing heading direction (no sharp turns). The duration of the recordings varies from 4 to 18 minutes for each trajectory, resulting in a total of 55 minutes for a single IMU and 275 minutes for the whole training dataset.
\\
The IMU data was divided to time windows and corresponding target values (GT). The target is based on pairs of GNSSS-RTK measurements (E and N position components) that were processed into distances. The corresponding equation for the GT distance between two successive position measurements is:
\begin{equation}\label{eq:gt_dist}
\centering
D_{\dot{\imath}}=\sqrt{(E_{\dot{\imath}+1}-E_{\dot{\imath}})^{2}+(N_{\dot{\imath}+1}-N_{\dot{\imath}})^{2}}
\end{equation}
where $D_{\dot{\imath}}$ is the target distance at epoch i and $E_{\dot{\imath}}$ and $N_{\dot{\imath}}$ are the east and north coordinates, respectively.
\\
The IMU data was segmented based on the time between two RTK measurements. Since the IMU operates at a frequency of 120 Hz and the RTK at 10 Hz, one RTK sample corresponds to twelve IMU samples. We used a window size of 24 IMU samples. This means that to calculate the target distance, we took every second RTK measurement at the start ($t_k$) and at the end ($t_{k+1}$) of the time window. Figure \ref{fig:win_size} shows a schematic description of the window size. Using overlap between time windows allowed us to utilize all the available RTK measurements and to enlarge the training dataset.
\begin{figure}
\centering
\includegraphics[width=0.6\linewidth]{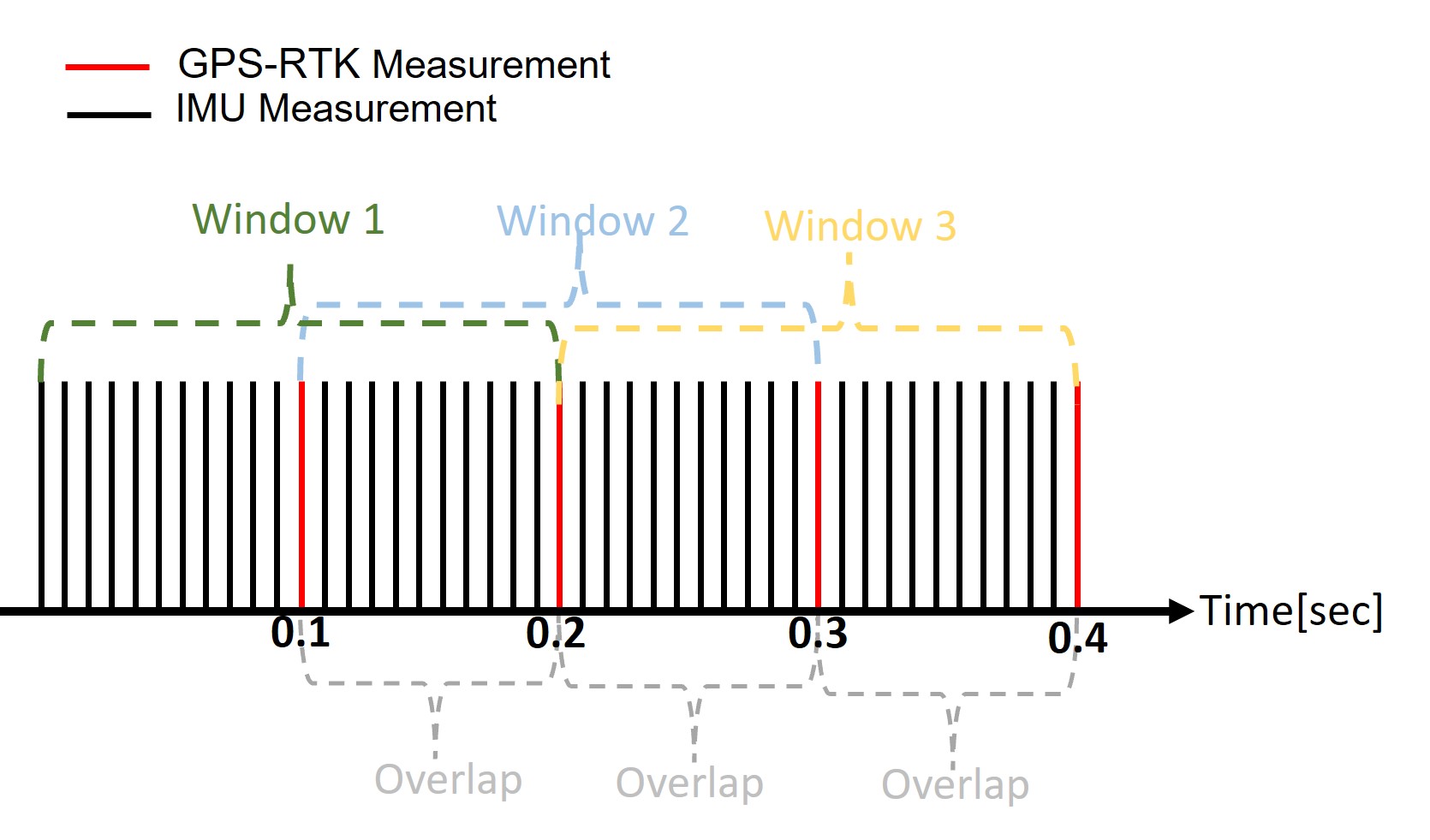}
\caption{Window size and overlap description }
\label{fig:win_size}
\end{figure}
\\
For MoRPI A and G methods, the role of the training dataset is to acquire the approach gain. According to \eqref{eq:weinberg}, the gain depends on the amplitude of the motion. Consequently, the gradual change in the trajectory's direction, reflected in varying amplitudes, affects the gain calculation. Thus, using the entire MoRPINet training dataset results in performance degradation. To address the above, sub-trajectories composed of straight segments from the MoRPINet training dataset are used to estimate the required MoRPI gain $G$ and lead to optimal results given this training dataset. These sub-trajectories are consistent with MoRPI, as MoRPI is specifically designed to operate on paths composed of straight segments. The MoRPI gain train dataset contains of five minutes of recordings.
\\
In addition, the MoRPI train dataset needs to be similar to the test dataset. However, the variance in step length across these trajectories is high, with a standard deviation of $1.49 m$, and the overall mean is $2.45 m$, compared to $2.02 m$ in the testing set. Consequently, MoRPI's performance is not ideal. The reason is that the dataset was recorded to demonstrate the robustness of the proposed approach and was not specifically tailored to the MoRPI method. Yet, such trajectories reflect real-world scenarios. Therefore, taking the above steps allows the model-based MoRPI approach to perform better.

\item \textbf{Test Dataset}:
The test dataset includes four trajectories of driving between two fixed points with a distance of about 25 meters. Those trajectories were recorded while the robot was moving in a snake-like slithering motion. The test dataset is the same for all baseline methods (INS and MoRPI) and the proposed MoRPINet approach. During assessment time, we tested MoRPI using the gain achieved by the corresponding training group, and for MoRPINet we used D-net with the optimal weights from the training process.
\\
The test group contains 3 minutes of recordings for a single IMU and 15 minutes for the whole test dataset.
\end{enumerate}
Additionally, two straight-line motion trajectories with a distance of about 25 meters were recorded. We used the two straight-line motion trajectories to evaluate only the INS method as the common baseline setup. The total time of these trajectories is approximately 4 minutes.
\\
In all the trajectories, initial heading angle was obtained by:

\begin{equation}\label{eq:gt_angs}
\centering
\psi_{\dot{\imath}}=atan2\left(\frac{N_{\dot{\imath}+1}-N_{\dot{\imath}}}{E_{\dot{\imath}+1}-E_{\dot{\imath}}}\right)
\end{equation}
where $\psi_{\dot{\imath}}$ is the true heading at epoch i. An illustration of the position, distance, and heading is provided in Figure \ref{fig:calc}.

\section{Analysis and Results}
In this section, we detail the evaluation metrics employed to assess the performance of our proposed method and present the results of the experiments conducted. Their relevance drove the selection of metrics to the specific objectives, the position error and the D-Net accuracy. The results highlight the comparative performance of our approach against baseline and alternative methods, emphasizing improvements and including visualizations to facilitate interpretation. 
\subsection{Evaluation Metrics} \label{subsec:EM}
To evaluate the performance of our approach, four metrics were used. Two for the position vector accuracy:
\begin{enumerate}
    \item \textbf{Position root mean squared error (PRMSE):}
    \begin{equation}\label{eq:prmse}
    \centering
        PRMSE(\boldsymbol{x}_{\dot\imath},\hat{\boldsymbol{x}}_{\dot\imath})=\sqrt{\frac{\sum_{\dot\imath=1}^{N}|\boldsymbol{x}_{\dot\imath}-\hat{\boldsymbol{x}}_{\dot\imath}|^{2}}{N}}
    \end{equation}
    
    \item \textbf{Position mean absolute error (PMAE):}
    \begin{equation}\label{eq:pmae}
    \centering
    PMAE(\boldsymbol{x}_{\dot\imath},\hat{\boldsymbol{x}}_{\dot\imath})=\frac{\sum_{\dot\imath=1}^{N}|\boldsymbol{x}_{\dot\imath}-\hat{\boldsymbol{x}}_{\dot\imath}|}{N}
    \end{equation}
\end{enumerate}
In \eqref{eq:prmse}-\eqref{eq:pmae} $\boldsymbol{x}_i$ is the measured position vector, $\hat{\boldsymbol{x}}_i$ is the expected position vector, and N is the number of samples.
\\
Two more metrics are used for the D-Net assessment and used to estimate the distance error of the step's distances as follows:
\begin{enumerate}[resume]
    \item \textbf{Distance root mean squared error (DRMSE):}
    \begin{equation}\label{eq:drmse}
    \centering
        DRMSE(d_{\dot\imath},\hat{d}_{\dot\imath})=\sqrt{\frac{\sum_{\dot\imath=1}^{N}\left(d_{\dot\imath}-\hat{d}_{\dot\imath}\right)^{2}}{N}}
    \end{equation}
    
    \item \textbf{Distance mean absolute error (DMAE):}
    \begin{equation}\label{eq:dmae}
    \centering
    DMAE(d_{\dot\imath},\hat{d}_{\dot\imath})=\frac{\sum_{\dot\imath=1}^{N}\left(d_{\dot\imath}-\hat{d}_{\dot\imath}\right)}{N}
    \end{equation}
\end{enumerate}
Where, in \eqref{eq:drmse}-\eqref{eq:dmae}, $d_i$ is the estimated distance from D-Net, $\hat{d}_i$ is the expected distance, and N is the number of steps.

\subsection{Field Experiment Results} \label{subsec:res}
MoRPINet, MoRPI and INS methods were evaluated for each test trajectory. Initially, the experiments and errors of the INS method were presented using the PRMSE. Then, a summary  of the D-Net performances is given. Finally, we compare the MoRPI and MoRPINet methods and analyze the results using the evaluation metrics \eqref{eq:prmse}-\eqref{eq:pmae}.
In the subsections below, we calculated the estimated position for each recorded IMU samples. The presented results are the average of the five recordings' position errors corresponding to the same trajectory (which were recorded simultaneously by the five mounted IMUs).

\subsubsection{Inertial Navigation System}
The INS solution was applied to the straight line trajectories for a fair comparison, as this the most common baseline navigation solution. Additionally, an analysis was conducted assuming ground planar movement, which means assuming 2D motion for the mobile robot. In \eqref{eq:ins_eq2} we presume $f_z=0$ and in \eqref{eq:ins_eq3} $\omega_x=\omega_y=0$. These assumptions were made to minimize system noise and achieve optimal results for this method \cite{etzion2023morpi}.
\\
We used three seconds with stationary conditions for each recording for calibration, as mentioned in subsection \ref{subsec:DS}. During this period, biases were extracted. This calibration technique was consistently applied across all reviewed methods when calibration was conducted. Specifically, calibration was performed for both accelerometers and gyroscopes in the INS solution.
\\
As expected, there is a significant error when using the INS equations. The PRMSE for the trajectories is $4502m$ using the 3D INS. Under the planar assumption, the 2D INS error is $295m$. The average distance of the trajectories is $24.4m$, and the mean travel time along the trajectories is $41s$. The errors for each trajectory are presented in Table \ref{tab:INSlines}.
\\
\begin{table}[!ht]
\caption{INS PRMSE results over the straight line trajectories in meters.}
\label{tab:INSlines}
\centering
\begin{tabular}{|c | c | c |}
\hline
Trajectory & 3D [m] & 2D [m] \\
\hline
1 & 3334 & 265 \\
\hline
2 & 5670 & 326 \\
\hline
avg & 4502 & 295 \\
\hline
\end{tabular}
\end{table}
\\
Additionally, we evaluated the INS solution for trajectories involving periodic movement from the test group. This type of movement results in a minor improvement compared to the straight-line movement.
\\
The average error of those trajectories for the 3D INS is $3528m$, and for the 2D INS is $262m$. The results are summarized in Table \ref{tab:INSpm}.
\begin{table}[!ht]
\caption{INS PRMSE results over the periodic movement test trajectories in meters.}
\label{tab:INSpm}
\centering
\begin{tabular}{|c | c | c |}
\hline
 Trajectory & 3D [m] & 2D [m]\\
\hline
3 & 2974 & 264\\
\hline
4 & 2876 & 220\\
\hline
5 & 3628 & 265\\
\hline
6 & 4633 & 298\\
\hline
avg & 3528 & 262\\
\hline
\end{tabular}
\end{table}

\subsubsection{D-Net}
 The IMU samples used as input to the neural network were not calibrated; however, calibration was part of the process in the AHRS filter.
    The suggested network was trained with the hyperparameters presented in Table \ref{table:hp} as they obtained the highest performance.
    \begin{table}[ht!]
    \caption{The hyper-parameters which were used in MoRPINet}
    \centering
    \begin{tabular}{|c|c|c|c|c|c|}
    \hline
    Learning rate & mini-batch & Epoch & Window & Overlap \\ \hline
    0.0025         & 2048       & 300   & 2      & 1           \\ \hline
    \end{tabular}
    \label{table:hp}
    \end{table}
    \\
Our D-Net achieved accuracy of 84\% and 87\% in terms of DRMSE and DMAE, respectively, relative to the average GT distance on the test set, as presented in Table \ref{tab:dnet}.
\begin{table}[ht!]
    \centering
    \begin{tabular}{|c|c|c|}
    \hline
      & DRMSE & DMAE \\
    \hline
    Error [m] & 0.022 & 0.017 \\
    \hline
    Accuracy [\%] & 83 & 87 \\
    \hline
    \end{tabular}
    \caption{MoRPINet D-Net errors in meters and percents relative to the GT distances.}
    \label{tab:dnet}
\end{table}

\subsubsection{MoRPI and MoRPINet}
The use of MoRPI requires obtaining the gain prior to position evaluation. We calculated the gain using straight segments cropped from the training group as described in Section \ref{subsec:DS}. The training set was constructed to ensure robust results from the neural network, containing trajectories with varying amplitudes and step sizes for the periodic movement. However, this characteristic does not align with MoRPI's requirements, where the training set should closely match the test set regarding amplitude and step size.
\\
The properties of the dataset are as follows: the average step size in the training set is $2.45 m$ with a standard deviation of $1.49$, while in the test set, the average step size and standard deviation are $2.02 m$ and $0.76$, respectively. Consequently, this mismatch led to a degradation in results, particularly in the MoRPI-G method.
\\
Despite this difficulty, MoRPI's results are significantly better than those of the INS method and are competitive with the proposed approach. Examining the test set, we observed an average error of $2.75 m$ using MoRPI-A with gyro calibration. For MoRPI-G, either with gyro calibration, the average error was $6.28 m$.
\\
The average number of position updates for the given test trajectories, with an average length of $24.6 m$, was $12.25$. The average time between two updates was $3.9 seconds$. The average travel time of a trajectory is 39 seconds. 
\\
 The results of MoRPINet show it scored a PRMSE of $1.92m$ on average of the test dataset, which is about $30\%$ improvement over MoRPI. The PMAE scores of MoRPINet was $1.59m$ averaged across the test trajectories. MoRPI, on the other hand, scored only $2.75m$ and $2.36m$ over the PRMSE and PMAE metrics, respectively. The results are summed in Table \ref{tab:morpi_res}.
    \begin{table}[!ht]
    \caption{MoRPI and MoRPINet PRMSE results in meters. The result of each trajectory in the table is the average of the PRMSE of the corresponding inertial reconstructed trajectories.}
    \label{tab:morpi_res}
    \centering
    \begin{tabular}{|c | c | c | c |}
    \hline
    Trajectory & MoRPI-A [m] & MoRPI-G [m] & MoRPINet [m]\\
    \hline
    3 & 4.23 & 3.16 & 2.06\\
    \hline    
    4 & 1.86 & 8.53 & 1.51\\
    \hline    
    5 & 3.09 & 6.44 & 2.16\\
    \hline     
    6 & 1.83 & 7.01 & 1.95\\
    \hline    
    avg & 2.75 & 6.28 & 1.92\\
    \hline
    \end{tabular}
    \end{table}
    \\
    \begin{figure*}[htb!]
    \begin{subfigure}[t]{0.5\linewidth}
    \includegraphics[width=\linewidth]{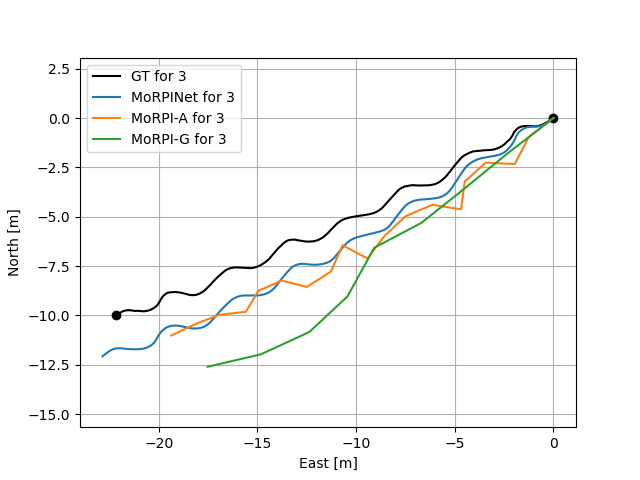}
        \caption{test trajectory no. 3}
    \end{subfigure}\hfill
    \begin{subfigure}[t]{0.5\linewidth}
    \includegraphics[width=\linewidth]{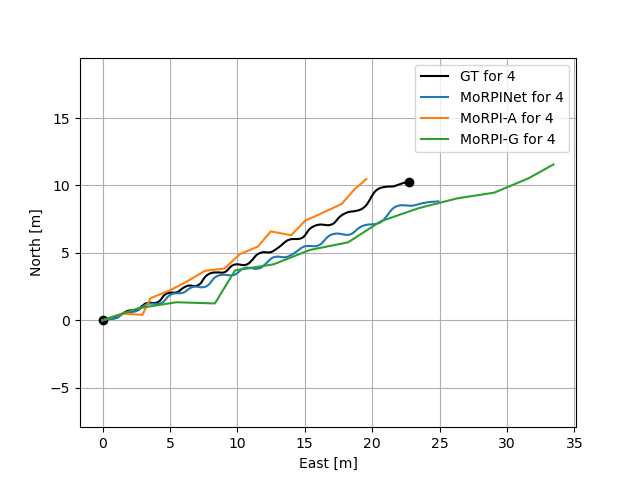}
        \caption{test trajectory no. 4}
    \end{subfigure}
    
    \begin{subfigure}[b]{0.5\linewidth}
    \includegraphics[width=\linewidth]{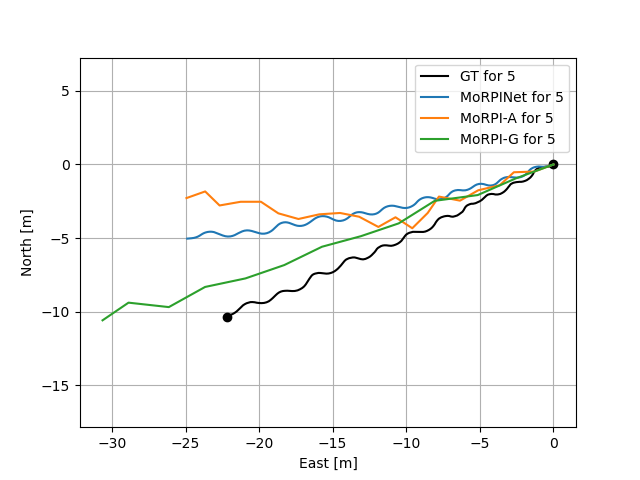}
        \caption{test trajectory no. 5}
    \end{subfigure}\hfill
    \begin{subfigure}[b]{0.5\linewidth}
        \includegraphics[width=\linewidth]{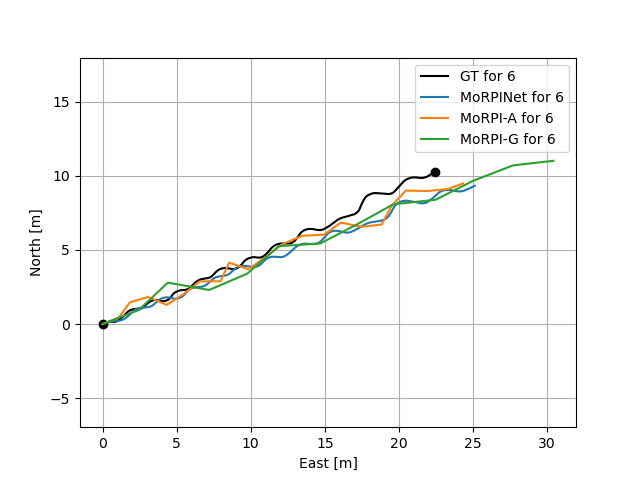}
        \caption{test trajectory no. 6}
    \end{subfigure}
    \caption{Comparison of horizontal position solution in different methods}
    \label{fig:res}
    \end{figure*}
   \\ 
    Table \ref{table:res} summarizes the results and presents a comparison between the different methods with the position evaluated metrics and update frequency. Additionally, Figure \ref{fig:res} presents a visual comparison of  the estimated trajectories from each approach and the associated GT trajectory. It is evident that our MoRPINet approach achieves the best positioning performance.
    \\
\begin{table}[ht!]
\caption{Comparison table of evaluation metrics averaged on four tested trajectories for each method. The MoRPI-A update rate is the average update rate due to the variance in the time windows as a dependency on the peaks.}
\centering
\begin{adjustbox}{width=\columnwidth}
\begin{tabular}{|l|c|cc|cc|}
\hline
\multicolumn{1}{|c|}{\multirow{2}{*}{}} & \multirow{2}{*}{\begin{tabular}[c]{@{}c@{}}update \\ rate {[}Hz{]}\end{tabular}} & \multicolumn{2}{c|}{PRMSE} & \multicolumn{2}{c|}{PMAE} \\ \cline{3-6} 
\multicolumn{1}{|c|}{} &  & \multicolumn{1}{c|}{{[}m{]}} & \begin{tabular}[c]{@{}c@{}}MoRPINet \\ improvement {[}\%{]}\end{tabular} & \multicolumn{1}{c|}{{[}m{]}} & \begin{tabular}[c]{@{}c@{}}MoRPINet \\ improvement {[}\%{]}\end{tabular} \\ \hline
MoRPINet (ours) & 5 & \multicolumn{1}{c|}{1.92} &  & \multicolumn{1}{c|}{1.59} &  \\ \hline
MoRPI-A & 0.26 & \multicolumn{1}{c|}{2.75} & 30 & \multicolumn{1}{c|}{2.36} & 33 \\ \hline
INS & 120 & \multicolumn{1}{c|}{262} & 99 & \multicolumn{1}{c|}{225} & 99  \\ \hline
\end{tabular}
\end{adjustbox}
\label{table:res}
\end{table}

\section{Conclusions} \label{sec:CON}
In real-world scenarios, during a mobile robot operation it commonly relies only on its inertial sensors for positioning. Yet, the error terms associated with the inertial readings cause the navigation solution to drift in time. To cope with such drift, we proposed the MoRPINet framework, an inertial data-driven approach that estimates the mobile robot position. MoRPINet framework consist of D-Net, a neural network architecture for distance regression, and a Madgwick filter for heading estimation. To increase the inertial signal to noise ratio, inspired by snakes’ movement, the mobile robot was maneuvered in serpentine locomotion allowing D-Net to accurately estimate the traveled distance.
\\
To evaluate our MoRPINet approach, five low-cost IMUs and an RTK-GNSS sensor were mounted on a mobile robot. The RTK position recordings, with 10cm accuracy, were used as GT. A dataset of 290 minutes (58 minutes from each IMU) of inertial recordings was used to train and test the model. We compared MoRPINet to two model-based approaches: 1) the commonly used INS solution and 2) the MoRPI approach which requires a gain calibration prior to its application. 
\\
 MoRPINet provides a position accuracy of only 1.59 meters for a 40 second and 24 meters long trajectory, according to PMAE. Compared to the MoRPI solution, the PMAE and the PRMSE along the RC car's route was reduced by 33\% and approximately 30\%, respectively. The INS solution rapidly drifts due to the accumulation of IMU errors and is practically irrelevant. The suggested D-Net achieved 87\% accuracy in terms of DMAE relative to the average GT step distance.
\\
The suggested approach presents three main advantages over the MoRPI approach: Improvement of at least 33\% over previous methods using a pure inertial solution. Furthermore, while MoRPI is a peak-to-peak-based method, the suggested approach is not limited and can be applied at a higher rate (for example 5Hz in our experiments which is 20 times faster than MoRPI), thus providing a better resolution. Finally, the MoRPI approach is highly sensitive to changes in the frequency of the periodic movement, while the suggested approach is more robust. It is also important to note that D-Net neural network is small, yet effective, and can be implemented on edge devices.
\\
In addition, all recorded data and code used for our evaluations are publicly available at \datasetUrl.
\\
In summary, MoRPINet offers an accurate, robust solution for mobile robot positioning in scenarios where only inertial reading are available. Future work could explore enhancing this framework to accommodate more challenging maneuvers, such as sharp turns or variations in elevation, further broadening its application potential.

\section*{Acknowledgment}
 A. E. and N. C. were supported by the Maurice Hatter Foundation

\bibliographystyle{IEEEtran}
\bibliography{./bib/references.bib}

\begin{thebibliography}{10}
\providecommand{\url}[1]{#1}
\csname url@samestyle\endcsname
\providecommand{\newblock}{\relax}
\providecommand{\bibinfo}[2]{#2}
\providecommand{\BIBentrySTDinterwordspacing}{\spaceskip=0pt\relax}
\providecommand{\BIBentryALTinterwordstretchfactor}{4}
\providecommand{\BIBentryALTinterwordspacing}{\spaceskip=\fontdimen2\font plus
\BIBentryALTinterwordstretchfactor\fontdimen3\font minus
  \fontdimen4\font\relax}
\providecommand{\BIBforeignlanguage}[2]{{%
\expandafter\ifx\csname l@#1\endcsname\relax
\typeout{** WARNING: IEEEtran.bst: No hyphenation pattern has been}%
\typeout{** loaded for the language `#1'. Using the pattern for}%
\typeout{** the default language instead.}%
\else
\language=\csname l@#1\endcsname
\fi
#2}}
\providecommand{\BIBdecl}{\relax}
\BIBdecl

\bibitem{rubio2019review}
F.~Rubio, F.~Valero, and C.~Llopis-Albert, ``A review of mobile robots:
  Concepts, methods, theoretical framework, and applications,''
  \emph{International Journal of Advanced Robotic Systems}, vol.~16, no.~2, p.
  1729881419839596, 2019.

\bibitem{vision2002}
G.~N. Desouza and A.~C. Kak, ``Vision for mobile robot navigation: A survey,''
  \emph{IEEE transactions on pattern analysis and machine intelligence},
  vol.~24, pp. 237--267, 2002.

\bibitem{lidar2018}
Y.~Cheng and G.~Y. Wang, ``Mobile robot navigation based on lidar,'' in
  \emph{2018 Chinese Control And Decision Conference ({CCDC})}, 2018, pp.
  1243--1246.

\bibitem{Leonard2012}
J.~J. Leonard and H.~F. Durrant-Whyte, \emph{Directed Sonar Sensing for Mobile
  Robot Navigation}.\hskip 1em plus 0.5em minus 0.4em\relax Springer Science \&
  Business Media, 2012, vol. 175.

\bibitem{farrell2008aided}
J.~Farrell, \emph{Aided navigation: GPS with high rate sensors}.\hskip 1em plus
  0.5em minus 0.4em\relax McGraw-Hill, Inc., 2008.

\bibitem{jimenez2010imu}
A.~R. Jim{\'e}nez, F.~Seco, J.~C. Prieto, and J.~Guevara, ``Indoor pedestrian
  navigation using an {INS/EKF} framework for yaw drift reduction and a
  foot-mounted {IMU},'' in \emph{2010 7th workshop on positioning, navigation
  and communication}.\hskip 1em plus 0.5em minus 0.4em\relax IEEE, 2010, pp.
  135--143.

\bibitem{Odometer2021}
D.~Nemec, M.~Hrubos, A.~Janota, R.~Pirnik, and M.~Gregor, ``Estimation of the
  speed from the odometer readings using optimized curve-fitting filter,''
  \emph{IEEE Sensors Journal}, vol.~21, no.~14, pp. 15\,687--15\,695, 2021.

\bibitem{kalman1960new}
R.~E. Kalman, ``{A New Approach to Linear Filtering and Prediction Problems},''
  \emph{Journal of Basic Engineering}, vol.~82, no.~1, pp. 35--45, 03 1960.

\bibitem{Titterton2005}
D.~H. Titterton and J.~L. Weston, \emph{\BIBforeignlanguage{eng}{Strapdown
  inertial navigation technology / D. H. Titterton, J. L. Weston.}}, 2nd~ed.,
  ser. Progress in astronautics and aeronautics vol. 207.\hskip 1em plus 0.5em
  minus 0.4em\relax Reston, Va: AIAA, 2004.

\bibitem{Itzik2022DGON}
I.~Klein, ``Data-driven meets navigation: Concepts, models, and experimental
  validation,'' in \emph{2022 {DGON} Inertial Sensors and Systems ({ISS})},
  2022, pp. 1--21.

\bibitem{cohen2023survey}
N.~Cohen and I.~Klein, ``Inertial navigation meets deep learning: A survey of
  current trends and future directions,'' \emph{arXiv preprint
  arXiv:2307.00014}, 2023.

\bibitem{chen2024survey}
C.~Chen and X.~Pan, ``Deep learning for inertial positioning: A survey,''
  \emph{IEEE transactions on intelligent transportation systems a publication
  of the IEEE Intelligent Transportation Systems Council.}, vol.~25, no.~9,
  2024-9.

\bibitem{yona2021compensating}
M.~Yona and I.~Klein, ``Compensating for partial {Dopple}r velocity log outages
  by using deep-learning approaches,'' in \emph{2021 IEEE International
  Symposium on Robotic and Sensors Environments (ROSE)}.\hskip 1em plus 0.5em
  minus 0.4em\relax IEEE, 2021, pp. 1--5.

\bibitem{saksvik2021auv}
I.~B. Saksvik, A.~Alcocer, and V.~Hassani, ``A deep learning approach to
  dead-reckoning navigation for autonomous underwater vehicles with limited
  sensor payloads,'' in \emph{OCEANS 2021: San Diego--Porto}.\hskip 1em plus
  0.5em minus 0.4em\relax IEEE, 2021, pp. 1--9.

\bibitem{shurin2022quadnet}
A.~Shurin and I.~Klein, ``{QuadNet}: A hybrid framework for quadrotor dead
  reckoning,'' \emph{Sensors}, vol.~22, no.~4, p. 1426, 2022.

\bibitem{zhang2022dido}
K.~Zhang, C.~Jiang, J.~Li, S.~Yang, T.~Ma, C.~Xu, and F.~Gao, ``Dido: Deep
  inertial quadrotor dynamical odometry,'' \emph{IEEE Robotics and Automation
  Letters}, vol.~7, no.~4, pp. 9083--9090, 2022.

\bibitem{asraf2021pdrnet}
O.~Asraf, F.~Shama, and I.~Klein, ``{PDRNet}: A deep-learning pedestrian dead
  reckoning framework,'' \emph{IEEE Sensors Journal}, vol.~22, no.~6, pp.
  4932--4939, 2021.

\bibitem{chen2020pdr}
C.~Chen, P.~Zhao, C.~X. Lu, W.~Wang, A.~Markham, and N.~Trigoni,
  ``Deep-learning-based pedestrian inertial navigation: Methods, data set, and
  on-device inference,'' \emph{IEEE Internet of Things Journal}, vol.~7, no.~5,
  pp. 4431--4441, 2020.

\bibitem{DOURADO2019859vision}
\BIBentryALTinterwordspacing
C.~M. Dourado, S.~P. {da Silva}, R.~V. {da Nóbrega}, A.~C. Barros, A.~K.
  Sangaiah, P.~P. {Rebouças Filho}, and V.~H.~C. {de Albuquerque}, ``A new
  approach for mobile robot localization based on an online {IoT} system,''
  \emph{Future Generation Computer Systems}, vol. 100, pp. 859--881, 2019.
  [Online]. Available:
  \url{https://www.sciencedirect.com/science/article/pii/S0167739X18317400}
\BIBentrySTDinterwordspacing

\bibitem{kim2015probabilisticVision}
H.~Kim, D.~Lee, T.~Oh, H.-T. Choi, and H.~Myung, ``A probabilistic feature
  map-based localization system using a monocular camera,'' \emph{Sensors},
  vol.~15, no.~9, pp. 21\,636--21\,659, 2015.

\bibitem{wang2019machineSlam}
X.~Wang, X.~Wang, and D.~M. Wilkes, \emph{Machine learning-based natural scene
  recognition for mobile robot localization in an unknown environment}.\hskip
  1em plus 0.5em minus 0.4em\relax Springer, 2019.

\bibitem{SLAM2004}
D.~Burschka and G.~D. Hager, ``{V-GPS(SLAM)}: Vision-based inertial system for
  mobile robots,'' \emph{Proceedings - IEEE International Conference on
  Robotics and Automation}, vol. 2004, pp. 409--415, 2004.

\bibitem{jayne1986kinematics}
B.~C. Jayne, ``Kinematics of terrestrial snake locomotion,'' \emph{Copeia}, pp.
  915--927, 1986.

\bibitem{hu2009mechanics}
D.~L. Hu, J.~Nirody, T.~Scott, and M.~J. Shelley, ``The mechanics of slithering
  locomotion,'' \emph{Proceedings of the National Academy of Sciences}, vol.
  106, no.~25, pp. 10\,081--10\,085, 2009.

\bibitem{Shurin2020}
A.~Shurin and I.~Klein, ``{QDR}: A quadrotor dead reckoning framework,''
  \emph{IEEE Access}, vol.~8, pp. 204\,433--204\,440, 2020.

\bibitem{hurwitz2023quadrotor}
D.~Hurwitz and I.~Klein, ``Quadrotor dead reckoning with multiple inertial
  sensors,'' in \emph{2023 DGON Inertial Sensors and Systems (ISS)}, 2023, pp.
  1--18.

\bibitem{etzion2023morpi}
A.~Etzion and I.~Klein, ``{MoRPI}: Mobile robot pure inertial navigation,''
  \emph{IEEE Journal of Indoor and Seamless Positioning and Navigation},
  vol.~1, pp. 141--150, 2023.

\bibitem{Groves2013}
P.~D. Groves, \emph{Principles of GNSS, Inertial and Multisensor Integrated
  Navigation Systems}, 2nd~ed.\hskip 1em plus 0.5em minus 0.4em\relax Artech
  House, 2013.

\bibitem{madgwick2010efficient}
S.~Madgwick \emph{et~al.}, ``An efficient orientation filter for inertial and
  inertial/magnetic sensor arrays,'' \emph{Report x-io and University of
  Bristol (UK)}, vol.~25, pp. 113--118, 2010.

\bibitem{NIPS2012relu}
\BIBentryALTinterwordspacing
A.~Krizhevsky, I.~Sutskever, and G.~E. Hinton, ``{ImageNet} classification with
  deep convolutional neural networks,'' in \emph{Advances in Neural Information
  Processing Systems}, F.~Pereira, C.~Burges, L.~Bottou, and K.~Weinberger,
  Eds., vol.~25.\hskip 1em plus 0.5em minus 0.4em\relax Curran Associates,
  Inc., 2012. [Online]. Available:
  \url{https://proceedings.neurips.cc/paper_files/paper/2012/file/c399862d3b9d6b76c8436e924a68c45b-Paper.pdf}
\BIBentrySTDinterwordspacing

\bibitem{bengio2017deep}
Y.~Bengio, I.~Goodfellow, and A.~Courville, \emph{Deep learning}.\hskip 1em
  plus 0.5em minus 0.4em\relax MIT press Cambridge, MA, USA, 2017, vol.~1.

\bibitem{gonzalez2018deep}
R.~C. Gonzalez, ``Deep convolutional neural networks [lecture notes],''
  \emph{IEEE Signal Processing Magazine}, vol.~35, no.~6, pp. 79--87, 2018.

\bibitem{agarap2018deep}
A.~F. Agarap, ``Deep learning using rectified linear units ({ReLU}),''
  \emph{arXiv preprint arXiv:1803.08375}, 2018.

\bibitem{zhao2017convolutional}
B.~Zhao, H.~Lu, S.~Chen, J.~Liu, and D.~Wu, ``Convolutional neural networks for
  time series classification,'' \emph{Journal of Systems Engineering and
  Electronics}, vol.~28, no.~1, pp. 162--169, 2017.

\bibitem{bock2019proof}
S.~Bock and M.~Wei{\ss}, ``{A proof of local convergence for the Adam
  optimizer},'' in \emph{2019 International Joint Conference on Neural Networks
  (IJCNN)}.\hskip 1em plus 0.5em minus 0.4em\relax IEEE, 2019, pp. 1--8.

\bibitem{Javad}
Javad, ``{Javad SIGMA-3N},'' Available:
  \url{https://www.javad.com/jgnss/products/receivers/sigma.html}, accessed:
  2022-10-01.

\bibitem{Xsens}
Xsens, ``{Xsens DOT},'' Available: \url{https://www.xsens.com/xsens-dot},
  accessed: 2022-10-01.

\end{thebibliography}

\end{document}